\documentclass[conference]{IEEEtran}
\IEEEoverridecommandlockouts
\usepackage{cite}
\usepackage{amsmath,amssymb,amsfonts}
\usepackage{algorithmic}
\usepackage{graphicx}
\usepackage{textcomp}
\usepackage{amsthm}
\usepackage{xcolor}
\usepackage{bbding}
\usepackage{makecell}
\usepackage{algorithm}  
\usepackage{color}
\usepackage{multirow}
\usepackage{bm}
\usepackage{array}
\usepackage{booktabs}
\usepackage{tabularx,booktabs}
\usepackage{multicol}
\usepackage{cleveref}
\usepackage{balance}
\usepackage{lastpage}
\usepackage{tabulary}
\usepackage{subfigure}
\usepackage{etoolbox}
\usepackage{bbm}
\usepackage{enumitem}
\usepackage{tcolorbox}
\usepackage{mathrsfs}
\usepackage{multicol}
\usepackage{amsfonts,amssymb}
\usepackage{ulem}
\usepackage{cancel}
\usepackage{setspace}
\usepackage{stackengine}
\usepackage{threeparttable}
\usepackage{fancyhdr}
\usepackage[letterpaper,
            bindingoffset=0in,
            left=0.635in,
            right=0.635in,
            top=0.75in,
            bottom=1.05in]{geometry}

\usepackage{amsthm}
\theoremstyle{definition}

\usepackage{nomencl}
\makenomenclature

\def\BibTeX{{\rm B\kern-.05em{\sc i\kern-.025em b}\kern-.08em
    T\kern-.1667em\lower.7ex\hbox{E}\kern-.125emX}}

\begin{document}

\bstctlcite{IEEEexample:BSTcontrol}

\title{Enhancing Large Language Models (LLMs) for Telecommunications using Knowledge Graphs and Retrieval-Augmented Generation

\author{\IEEEauthorblockN{ Dun Yuan$^1$, Hao Zhou$^1$, Di Wu$^2$, Xue Liu$^1$, \IEEEmembership{Fellow, IEEE}, \\
Hao Chen$^3$, Yan Xin$^3$,  Jianzhong (Charlie) Zhang$^3$, \IEEEmembership{Fellow, IEEE} }
\IEEEauthorblockA{\textit{School of Computer Science$^1$/ Department of Electrical and Computer Engineering$^2$, McGill University}\\ 
\textit{Standards and Mobility Innovation Lab, Samsung Research America$^3$}\\
Emails:\{dun.yuan, hao.zhou4\}@mail.mcgill.ca, xueliu@cs.mcgill.ca, di.wu5@mcgill.ca}
\{hao.chen1, yan.xin, jianzhong.z\}@samsung.com}
\vspace{-5pt}}

\maketitle

\begin{abstract}
Large language models (LLMs) have made significant progress in general-purpose natural language processing tasks. However, LLMs are still facing challenges when applied to domain-specific areas like telecommunications, which demands specialized expertise and adaptability to evolving standards. 
This paper presents a novel framework that combines knowledge graph (KG) and retrieval-augmented generation (RAG) techniques to enhance LLM performance in the telecom domain. The framework leverages a KG to capture structured, domain-specific information about network protocols, standards, and other telecom-related entities, comprehensively representing their relationships.
By integrating KG with RAG, LLMs can dynamically access and utilize the most relevant and up-to-date knowledge during response generation. This hybrid approach bridges the gap between structured knowledge representation and the generative capabilities of LLMs, significantly enhancing accuracy, adaptability, and domain-specific comprehension. Our results demonstrate the effectiveness of the KG-RAG framework in addressing complex technical queries with precision.
The proposed KG-RAG model attained an accuracy of 88\% for question answering tasks on a frequently used telecom-specific dataset, compared to 82\% for the RAG-only and 48\% for the LLM-only approaches.

\end{abstract}


\begin{figure*}[t]
\centering
\includegraphics[width=0.7\linewidth]{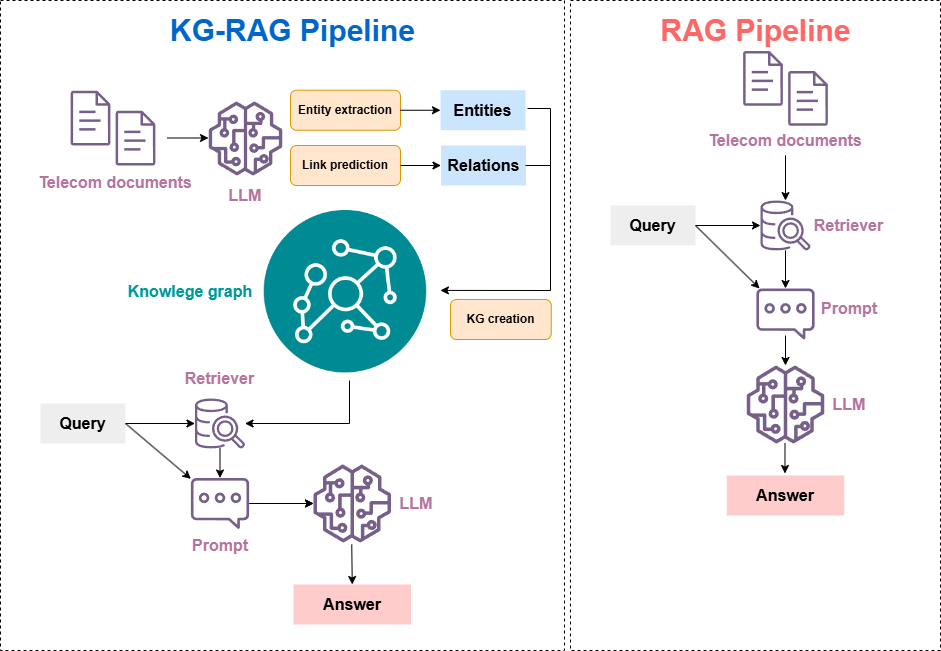}
\caption{Illustration of the KG-RAG pipeline compared to the traditional RAG pipeline. Unlike the common RAG pipeline, the KG-RAG approach leverages a KG as a comprehensive knowledge base, providing a more structured and dynamic response capability.}
\label{fig:method}
\end{figure*}

\section{Introduction}
Large language models (LLMs) have exhibited remarkable capabilities across various natural language processing tasks, including text generation, question answering, and knowledge retrieval\cite{edge2024local}. 
However, these models often underperform in domain-specific applications, particularly in highly specialized fields like telecom fields\cite{zhou2024large}. 
This limitation arises because traditional LLMs are predominantly trained on general-purpose data sources. Consequently, these models lack the specialized understanding required to navigate the complexities of telecom-related queries, which demand in-depth knowledge of cellular network protocols, industry standards, and rapidly evolving communication technologies.
In addition, the telecom industry is characterized by intricate systems and swiftly updating standards set by organizations such as the 3rd Generation Partnership Project (3GPP) and the Open Radio Access Network (ORAN) Alliance. 
Developing an LLM that can effectively comprehend and reason about telecom knowledge indicates great benefits for the telecom industry, which can be further applied to network troubleshooting, project coding, standard development, etc.     

Enabling general LLMs to understand telecom-domain knowledge is a significant challenge. A widely considered approach is to fine-tune LLMs on specific telecom datasets. 
However, there are several potential challenges for fine-tuning large language models on domain-specific data, including high computational costs and inefficiency in capturing the rapidly evolving nature of telecom standards. Fine-tuning requires substantial computational resources and time, especially for very large models, making it impractical for frequent updates. In addition, fine-tuned models may become outdated quickly as new standards and protocols are developed, and there is a risk of overfitting the fine-tuning data, leading to a loss of generalization capability.
Given the great potential of LLMs and the challenges in telecom domain applications, this work proposes a novel approach to enhance LLMs for the telecom domain by integrating Retrieval-Augmented Generation (RAG) and Knowledge Graph (KG) techniques.

In particular, RAG refers to the process of enhancing language models by incorporating relevant information retrieved from external knowledge sources during response generation\cite{lewis2020retrieval}. This allows the model to access and utilize up-to-date and domain-specific information that is not captured in its static training data. The RAG framework can improve an LLM's accuracy by enabling it to dynamically retrieve relevant, current information from the KG during response generation.
RAG has emerged as a powerful approach to augment LLMs with domain knowledge. For instance, the study by Roychowdhury \textit{et al.}~\cite{roychowdhury2024evaluation} evaluates the use of RAG metrics for question answering in telecom applications, while Medical Graph RAG~\cite{wu2024medical} explores the use of graph-based retrieval for enhancing medical LLMs' safety. The authors of \cite{zitian2024enhancing} demonstrate the effectiveness of GraphRAG for improving predictions in venture capital.

Meanwhile, KG refers to the data structures that represent information in the form of entities and the relationships between them. They are used to model complex systems and domains by capturing the interconnections among various components. In the telecom domain, KG offers a structured representation of telecom-specific information by capturing relationships among key entities such as network protocols, signalling standards, and hardware components.
The integration of LLMs with KGs has also received significant attention. Research such as KG-LLM~\cite{shu2024knowledge} investigates the role of KGs in link prediction tasks, while Li \textit{et al.}~\cite{li2024enhanced} demonstrate how KG assists LLMs in complex reasoning tasks. Sun \textit{et al.}~\cite{sun2023head} question whether LLMs could replace KGs, proposing experiments to compare the two paradigms. Complementarily, Zhang, and Soh~\cite{zhang2024extract} propose a framework for using LLMs to construct KGs through extraction, definition, and canonicalization.

Combining KG with RAG indicates great potential to adapt LLMs to telecom domains, contributing to generative AI-enabled 6G networks.
The core contribution of this work is that we proposed a novel KG-RAG approach to enhance the knowledge understanding performance for LLMs in the telecom domain. 
Such a hybrid methodology leverages the strengths of structured knowledge representation and the generative capabilities of LLMs, resulting in a model that is more accurate and better suited to adapt to the evolving nature of telecom knowledge.
By employing RAG and KG, we aim to advance the application of LLMs in domain-specific contexts, using telecom as a primary example. 

Our proposed KG-RAG framework demonstrates significant improvements in domain-specific performance for telecommunications. For question answering tasks on the Tspec-LLM\cite{nikbakht2024tspec} dataset, the KG-RAG model achieved an accuracy of 88\%, compared to 82\% for the RAG model and 48\% for the LLM-only approach. These results show the effectiveness of the KG-RAG model that enables LLMs to better handle technical queries with precision and up-to-date knowledge.

The rest of this paper is organized as follows: Section II introduces the overall system model and provides a technical background on KGs. Section III presents the KG-RAG framework, detailing the retrieval mechanism, contextual retrieval, and generation processes. In Section IV, we describe our experimental setup, datasets, and evaluation metrics, followed by the result tables and figures in text summarization and question answering tasks. Finally, Section V concludes the paper with a discussion of key findings, potential applications, and potential future work directions.

\section{System Model and Technical Background}

\subsection{Overall System Model}
The proposed system model leverages the integration of LLMs, KGs, and RAG frameworks to enhance domain-specific language processing within the telecom field. As shown in Figure~\ref{fig:method}, the KG-RAG architecture is organized into several key components. 
The system starts by building a domain-specific KG that encapsulates key entities, such as network protocols, hardware components, and signalling standards, as well as their relationships. The KG is constructed using structured and unstructured data sources from the telecom domain, transforming these data points into triples of the form (head entity, relationship, tail entity).
When a query is presented to the system, the retrieval mechanism dynamically retrieves relevant subgraphs or nodes from the KG. This step uses similarity measures and relevance scoring to ensure that only the most pertinent knowledge is selected for further processing. The retrieved information provides a robust contextual base for subsequent operations.

Finally, the retrieved knowledge is incorporated into an LLM using a retrieval-augmented generation approach. This process involves conditioning the LLM's response on the retrieved knowledge, allowing the system to generate accurate and contextually appropriate responses to user queries. The LLM's generative capabilities are enhanced by the structured, domain-specific information from the KG, resulting in responses that are both precise and adaptable to evolving telecom standards. 
This hierarchical structure ensures that domain-specific knowledge is accurately captured, efficiently retrieved, and effectively used to enhance LLM responses.

In the following, we will introduce the knowledge graph, LLM-aided entity extraction, and link prediction.

\subsection{Knowledge Graph}
A KG is a structured representation of knowledge consisting of entities and their relationships. It is commonly used to model complex systems of information in fields such as natural language processing and semantic web.

\subsubsection{\textbf{Entities and Relationships}}

KG includes the following elements: $E$ as the set of all entities (nodes) in the KG, $R$ as the set of all relationships (edges) between entities, and $L$ as the set of all literals (attribute values).

\subsubsection{\textbf{Triples}}
The fundamental building block of a KG is a triple $(h,r,t)$, where $h\in E$ is the head entity, $r\in R$ is the relationship, and $t\in E\cup L$ is the tail, which can be either another entity or a literal. Thus, a KG can be formally defined as a set of triples: 
\begin{equation}
  KG=\{(h,r,t)\mid h\in E, r\in R, t\in E\cup L\}  
\end{equation}

\subsubsection{\textbf{Graph Representation}}
A KG can be visualized as a directed labelled graph $G=(V,E)$, where $V=E\cup L$ is the set of vertices (entities and literals), and $E$ is the set of directed edges representing relationships $r\in R$ between subjects and objects.
Each edge $e\in E$ can be represented as a tuple $e=(h,r,t)$.

\subsubsection{\textbf{Knowledge Graph Embeddings}}

To enable machine learning algorithms to process KGs, entities and relationships can be embedded into continuous vector spaces.
An embedding model maps entities and relationships to vectors as $f_E: E\rightarrow \mathbb{R}^d$ and $f_R: R\rightarrow \mathbb{R}^d$, where $d$ is the dimension of the embedding space.

Additionally, scoring functions are used to measure the plausibility of a triple $(h,r,t)$, e.g., the scoring function of TransE \cite{bordes2013translating} model is defined as
\begin{equation}
\text{score}(h,r,t)=-\|f_E(h)+f_R(r)-f_E(t)\|_2^2   
\end{equation}
where $f_E(h)$ and $f_E(t)$ represent the embeddings of the head entity $h$ and tail entity $t$, respectively, and $f_R(r)$ denotes the embedding of the relation $r$.

\subsection{LLM-aided Entity extraction}
Entity extraction is an important step in constructing a KG for a telecom-specific language model. It involves identifying key entities and their relationships from a variety of unstructured and semi-structured data sources. In the telecom domain, these entities include concepts like network protocols, hardware components, signal types, frequency bands, and communication standards, among others. 
Accurate extraction of these entities enables the creation of a structured KG that serves as the backbone for enhancing the language model’s capabilities.

We employ the LLM to extract entities from input documents \(D=\{d_1,d_2,\ldots,d_m\}\). Each document is tokenized into smaller segments using predefined prompts to detect named entities and relationships. Given an input document \(d_i\), the LLM generates a set of entities \(E=\{e_1,e_2,\ldots,e_n\}\). Each entity \(e_i\) is assigned additional metadata \(M_i\), including type (e.g., protocol, metric, component) and semantic context.

The extraction process is formulated as
\begin{equation}
E=\text{LLM}_{\text{extract}}(D)
\end{equation}
where \(\text{LLM}_{\text{extract}}\) is the extraction module of the LLM, operating on the entire document set \(D\). Each entity \(e_i\) may also exhibit multiple relationships with other entities, such that
\begin{equation}
R=\{(e_i,e_j,r_{ij}) \mid e_i,e_j\in E, r_{ij} \in \mathcal{R}\}
\end{equation}
where \(\mathcal{R}\) represents the set of possible relationships.

\subsection{Link Prediction}
Link prediction plays a crucial role in constructing and expanding a KG where entities such as protocols, network components, and standards are highly interconnected. The task involves identifying potential relationships between entities that are either missing or can be inferred from existing connections within the KG.

To achieve this, link prediction models are trained to maximize the scores of true triples while minimizing the scores of false triples. This ensures that the models accurately predict valid connections, improving the overall utility and completeness of the KG.
The training objective often involves a margin-based ranking loss:
\begin{equation}
\mathcal{L}=\sum_{(h,r,t)\in \mathcal{T}} \sum_{(h',r',t')\in \mathcal{T}'}\left[ \gamma + f(h',r',t') - f(h,r,t) \right]_+
\end{equation}
where $\mathcal{T}$ is the set of positive triples, $\mathcal{T}'$ is the set of negative samples, $\gamma$ is the margin, and $[x]+ = \max(0,x)$ is the hinge loss. This encourages true triples to have scores at least $\gamma$ higher than false ones.
Negative samples are typically generated by corrupting true triples (e.g., replacing the tail entity), helping the model learn to differentiate valid from invalid facts.

\section{KG-RAG Framework}
This section introduces the KG-RAG framework, developed to enhance LLMs for domain-specific applications in telecommunications. By combining KG with RAG, the framework aims to improve the accuracy and relevance of LLM-generated responses. We provide details on the retrieval and generation components that form the core of the KG-RAG framework.

\subsubsection{Retrieval}
In the KG-RAG framework, the retrieval component is responsible for fetching relevant information from the KG to support the language model's response generation.

Let $\mathcal{Q}$ represent the set of all possible queries, and let $q \in \mathcal{Q}$ be a specific user query. The retrieval function $\mathcal{R}$ maps this query to a subset of the KG $\mathcal{K}$
\begin{equation}
\mathcal{R}: \mathcal{Q} \rightarrow 2^{\mathcal{K}}    
\end{equation}
where $2^{\mathcal{K}}$ denotes the power set of $\mathcal{K}$. The retrieval process involves the following steps:

\begin{enumerate}
    \item \textbf{Query Encoding}: Convert the user query $q$ into a vector representation $\mathbf{q} \in \mathbb{R}^d$ using an encoding function $\phi$:
    \begin{equation}
    \mathbf{q} = \phi(q)
    \end{equation}
    \item \textbf{Knowledge Encoding}: Encode all candidate knowledge snippets $k \in \mathcal{K}$ into vector representations $\mathbf{k} \in \mathbb{R}^d$.
    \item \textbf{Similarity Computation}: Compute the similarity between the query vector $\mathbf{q}$ and each knowledge vector $\mathbf{k}$ using a similarity measure such as cosine similarity:
    \begin{equation}
    \text{sim}(\mathbf{q}, \mathbf{k}) = \frac{\mathbf{q} \cdot \mathbf{k}}{\|\mathbf{q}\| \|\mathbf{k}\|}
    \end{equation}
    \item \textbf{Retrieval of Top-$K$ Results}: Select the top-$K$ knowledge snippets that have the highest similarity scores with the query:
    \begin{equation}
    \mathcal{S}_q = \operatorname{arg\,max}_{k \in \mathcal{K}}^K \text{sim}(\mathbf{q}, \mathbf{k})
    \end{equation}
\end{enumerate}

The retrieval component effectively narrows down the vast knowledge base to a manageable and relevant subset $\mathcal{S}_q$ that can be used by the generation component.

The relevance of each knowledge snippet to the query is quantified using a scoring function $s(q, k)$, which may incorporate additional factors like term frequency-inverse document frequency (TF-IDF) weights, entity matching, or semantic similarity:
\begin{equation}
s(q, k) = \alpha \cdot \text{sim}(\mathbf{q}, \mathbf{k}) + \beta \cdot \text{TF-IDF}(q, k) + \gamma \cdot \text{EM}(q, k)   
\end{equation}
where $\alpha$, $\beta$, and $\gamma$ are weighting coefficients, and $\text{EM}(q, k)$ is an entity matching score.

\subsubsection{Generation}

The generation component uses the retrieved knowledge snippets $\mathcal{S}_q$ to produce a coherent and contextually appropriate response $r$ to the user query $q$.

The input to the language model combines the user query and the retrieved knowledge:
\begin{equation}
I = \text{Format}(q, \mathcal{S}_q)
\end{equation}
where $\text{Format}$ is a function that structures the input appropriately, such as by appending the knowledge snippets to the query with special tokens to delineate them.

The language model generates the response by maximizing the conditional probability:
\begin{equation}
P(r \mid I) = \prod_{t=1}^{T} P(r_t \mid r_{<t}, I)
\end{equation}
where $r_t$ is the $t$-th token in the response, $r_{<t}$ are the tokens generated so far, and $T$ is the total number of tokens in the response.

During generation, attention mechanisms allow the model to focus on relevant parts of the input, including both the query and the retrieved knowledge. The attention weight $\alpha_{t,i}$ for the $i$-th input token at time $t$ is computed as:
\begin{equation}
\alpha_{t,i} = \frac{\exp(e_{t,i})}{\sum_j \exp(e_{t,j})}
\end{equation}
\begin{equation}
e_{t,i} = \mathbf{h}_{t-1}^\top \mathbf{W}_a \mathbf{e}_i
\end{equation}
where $\mathbf{h}_{t-1}$ is the hidden state of the decoder at time $t-1$, $\mathbf{W}_a$ is a learned parameter matrix, and $\mathbf{e}_i$ is the embedding of the $i$-th input token.

The model is trained to minimize the negative log-likelihood loss:
\begin{equation}
\mathcal{L} = -\sum_{t=1}^{T} \log P(r_t \mid r_{<t}, I)
\end{equation}
To encourage the model to use the retrieved knowledge, an auxiliary loss term can be added to penalize deviations from the knowledge:
\begin{equation}
\mathcal{L}_{\text{total}} = \mathcal{L} + \lambda \cdot \mathcal{L}_{\text{knowledge}}
\end{equation}
where $\lambda$ is a hyperparameter balancing the two loss terms, and $\mathcal{L}_{\text{knowledge}}$ measures the discrepancy between the generated response and the retrieved knowledge.

To summarize, for any given query, the processes in KG-RAG ensure that the LLM has access to the most relevant domain-specific knowledge from the KG. The steps are summarized as follows:

\begin{itemize}
    \item \textbf{Query Encoding:} Encode the query into a semantic vector.
    \item \textbf{Graph Retrieval:} Retrieve nodes and edges from the KG using similarity matching.
    \item \textbf{Response Generation:} Use the retrieved subgraph as input to the LLM to generate a response.
\end{itemize}

The response generation step ensures that the KG-RAG framework can provide accurate, grounded answers by leveraging the structured knowledge from the KG.

\section{Performance Evaluation}

\subsection{Experiment Settings}

In this work, we use OpenAI's GPT-4o-mini \cite{4o-mini} model as the basic reasoning LLM for our experiments. This model demonstrates strong capabilities in reasoning tasks, including addressing complex queries such as telecom-related questions and broader analytical challenges. Its integration into our framework aligns with the nature of the datasets applied in our work, providing robust and contextually aware responses. Moreover, its cost-efficiency is particularly advantageous for analyzing and performing inference on large corpus datasets such as telecom documents.

\subsubsection{\textbf{Telecom Datasets}} 
\hfill \\
\textbf{\textcircled{1}: SPEC5G}\cite{karim2023spec5g}: It contains 134 million words and is a comprehensive collection of technical specifications and documentation related to 5G wireless technology. It includes standards from organizations such as 3GPP, ITU, and ETSI.\\
\textbf{\textcircled{2}: Tspec-LLM}\cite{nikbakht2024tspec}: Comprising 534 million words and 100 questions, Tspec-LLM is a specialized corpus of technical telecom documents curated for training and evaluating large language models.\\
\textbf{\textcircled{3}: TeleQnA}\cite{maatouk2023teleqna}: This dataset features 10,000 curated question-and-answer pairs focused on the telecom domain, providing a valuable resource for evaluation purposes. In order to utilize the ability of the framework, we use the documents of Tspec-LLM dataset as the knowledge database when answering questions.\\
\textbf{\textcircled{4}: ORAN-Bench-13K}\cite{gajjar2024oran}: This dataset consists of 2.53 million words and 13,000 multiple-choice questions generated from 116 O-RAN specification documents. The questions are categorized into three difficulty levels.

\subsubsection{\textbf{Evaluation Tasks}} \hfill \\
\textbf{\textcircled{1}: Text Summarization}: This task involves generating concise summaries of technical telecom documents to extract key information and insights. It helps in digesting large volumes of data by highlighting the most important aspects.\\
\textbf{\textcircled{2}: Question Answering}: This task focuses on providing accurate answers to queries based on telecom-related data. It evaluates the model's ability to understand questions and retrieve relevant information.

\subsubsection{\textbf{Baselines}} \hfill \\
\textbf{\textcircled{1}: LLM-only}: The LLM directly processes queries without external retrieval or structured knowledge. It relies solely on its internal knowledge.\\
\textbf{\textcircled{2}: RAG}: The RAG framework augments the LLM with a retrieval component, fetching relevant documents or snippets from datasets. The retrieved content provides additional context for response generation.\\
\textbf{\textcircled{3}: KG-RAG}: The introduced method in this work integrates KGs and the RAG pipeline.

\subsection{Results}

This subsection will present the evaluation results of our KG-RAG framework on the text summarization and question answering tasks. We compare the performance of three models: LLM-only, RAG, and our proposed KG-RAG model.

\subsubsection{Text summarization}

The text summarization task involves generating concise and informative summaries of technical telecom documents from the SPEC5G\cite{karim2023spec5g} dataset. The goal is to condense complex and lengthy technical specifications into shorter texts that retain the most important information, making them more accessible for analysis and decision-making.

We evaluate the performance of the models using the Recall-Oriented Understudy for Gisting Evaluation (ROUGE) metrics\cite{lin2004rouge}, which measure the overlap between the generated summaries and the reference summaries. Specifically, we use the ROUGE metrics. The ROUGE score is well-suited for this task because it captures the content similarity between the generated and reference summaries, which is critical in assessing the quality of summarization. ROUGE metrics used in this work include:
\begin{itemize}
    \item \textbf{ROUGE-1}: Measures the overlap of unigrams between the candidate and reference texts:
    \[
    \text{ROUGE-1} = \frac{\textit{Number of overlapping unigrams}}{\textit{Total unigrams in the reference}}
    \]
    \item \textbf{ROUGE-2}: Measures the overlap of bigrams between the candidate and reference texts:
    \[
    \text{ROUGE-2} = \frac{\textit{Number of overlapping bigrams}}{\textit{Total bigrams in the reference}}
    \]
    \item \textbf{ROUGE-L}: Measures the longest common subsequence between the candidate and reference texts, providing a measure of sequence-level similarity and capturing sentence structure:
    \[
    \text{ROUGE-L} = \frac{\textit{Length of LCS}}{\textit{Total words in the reference}}
    \]
\end{itemize}

\begin{table}[b]
\caption{Results of text summarization task on SPEC5G dataset}
\centering
\begin{tabular}{|c|c|c|c|c|}
\hline
Models& ROUGE-1&ROUGE-2&ROUGE-L\\ \hline
LLM-only    &0.53$\pm$0.02    &0.31$\pm$0.02    &0.44$\pm$0.03   \\ \hline
RAG         &0.55$\pm$0.03    &0.34$\pm$0.02    &0.45$\pm$0.02  \\ \hline
KG-RAG      &\textbf{0.58$\pm$0.02}    &\textbf{0.38$\pm$0.03}    &\textbf{0.46$\pm$0.02}  \\ \hline
\end{tabular}
\label{tab:results1}
\end{table}

\begin{table}[b]
\caption{Accuracy of each model on different datasets}
\centering
\begin{tabular}{|c|c|c|c|}
\hline
Models& Tspec-LLM& TeleQnA& ORAN-Bench-13K  \\ \hline
LLM-only    &0.48   &0.72   &0.26   \\ \hline
RAG         &0.82   &0.74   &0.72     \\ \hline
KG-RAG      &\textbf{0.88}   &\textbf{0.75}   &\textbf{0.80}     \\ \hline
\end{tabular}
\label{tab:accuracy}
\end{table}

Table~\ref{tab:results1} presents the mean and variance scores for all items in the SPEC5G\cite{karim2023spec5g} dataset. The results show that the KG-RAG model consistently outperforms both the LLM-only and RAG models across all ROUGE metrics. Notably, the KG-RAG model achieves a ROUGE-1 score of 0.58, which is a significant improvement over the LLM-only model's score of 0.53 and the RAG model's score of 0.55. Similar improvements can be observed in ROUGE-2 and ROUGE-L scores, indicating that the KG-RAG model generates summaries that are more similar to the reference summaries in terms of both content and sequence.

\subsubsection{Question answering}
We assess the effectiveness of each method in accurately answering telecom-specific questions. The evaluation metrics used include accuracy and response time.
To evaluate the performance of each model, we utilized sets of telecom-specific questions similar to the one above. The questions were designed to test the models' understanding of key concepts in telecom such as signal processing, network architecture, and wireless communication protocols.

As shown in Table~\ref{tab:accuracy}, the KG-RAG model consistently outperforms the LLM-only and RAG models across all datasets. Specifically, on the Tspec-LLM\cite{nikbakht2024tspec} dataset, the KG-RAG model achieves an accuracy of 0.88, a substantial improvement over the LLM-only's accuracy of 0.48 and the RAG's accuracy of 0.82. Similar performance gains are observed on the ORAN-Bench-13K\cite{gajjar2024oran} datasets. It is worth noting that the TeleQnA\cite{maatouk2023teleqna} datasets feature more general questions that are not specifically tied to particular documents. Consequently, the LLM-only approach, even without the RAG component, does not perform as poorly compared to other methods in this context.

\begin{figure}[t]
\centering
\includegraphics[width=\linewidth]{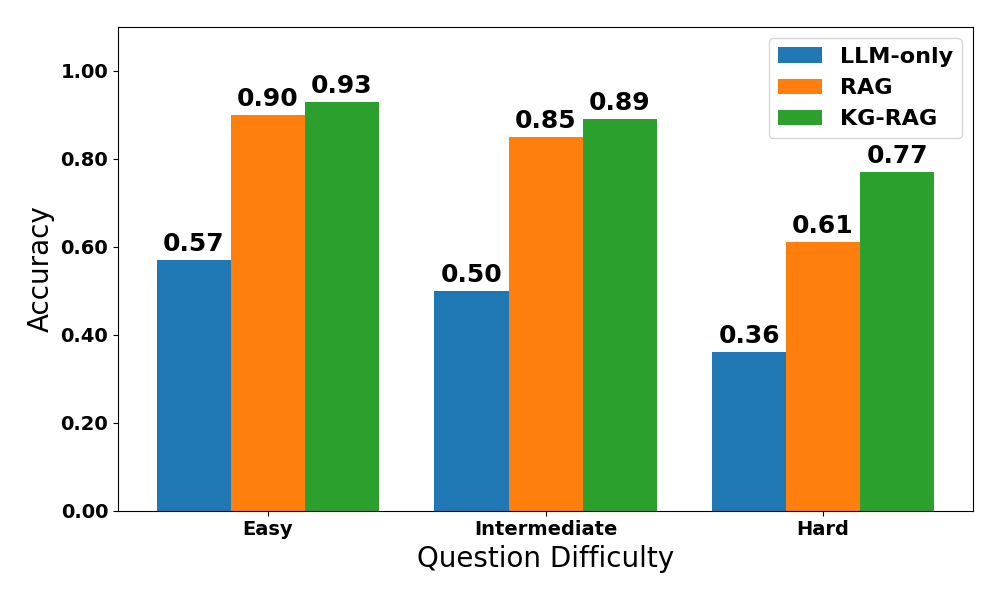}
\vspace{-20pt}
\caption{Accuarcies of different models by question difficulty on Tspec-LLM\cite{nikbakht2024tspec}}
\vspace{-5pt}
\label{fig:tspec}
\end{figure}

\begin{figure}[t]
\centering
\includegraphics[width=\linewidth]{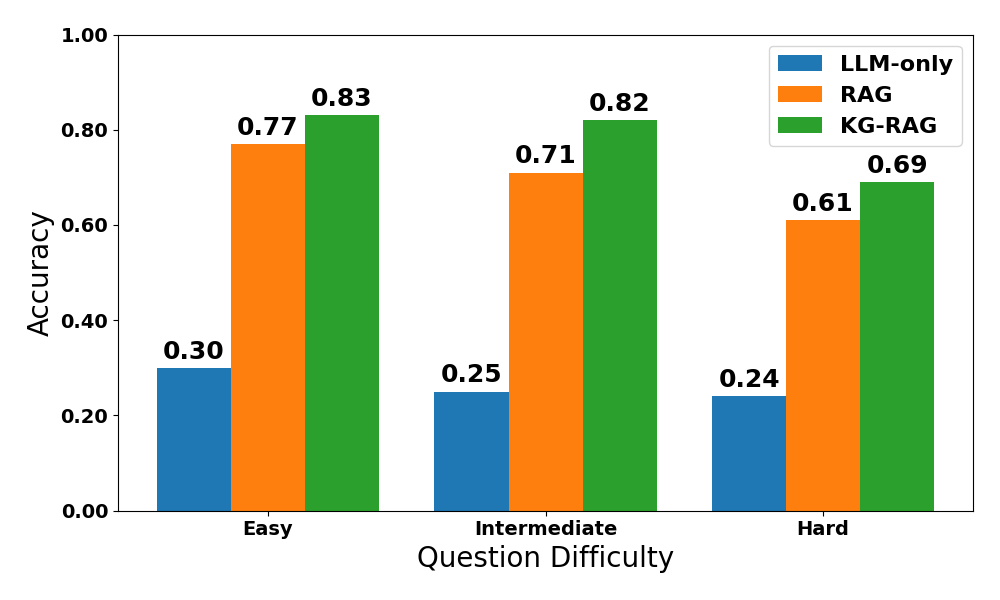}
\vspace{-20pt}
\caption{Accuarcies of different models by question difficulty on ORAN-Bench-13K\cite{gajjar2024oran}}
\label{fig:oran}

\end{figure}

We also examined the accuracy by question difficulty levels on the Tspec-LLM and ORAN-Bench-13K datasets. The dataset categorizes questions into three difficulty levels: easy, intermediate, and hard. From Figure~\ref{fig:tspec} and Figure~\ref{fig:oran}, we observe that the KG-RAG model demonstrates superior performance across all difficulty levels. The improvement is particularly notable for intermediate and hard questions, where domain-specific knowledge and reasoning are critical.

\section{Conclusion}

LLM is a promising technique for envisioned 6G networks. 
This paper introduced the KG-RAG framework, which integrates KG with RAG to enhance LLMs for the telecommunications domain. Our methodology addresses the limitations of general-domain LLMs by grounding responses in structured knowledge extracted from telecom-specific datasets. Our experiments demonstrated that KG-RAG outperforms LLM-only and traditional RAG models, showing significant gains across telecom datasets for question answering and summarization tasks.
Our work highlights the potential of integrating KGs with RAG framework to create domain-specific models that can effectively address the challenges of complex technical queries, providing a significant advancement in the application of LLMs within the telecom industry. In future work, we plan to explore the dynamic expansion of the KG by incorporating real-time updates from telecom industry standards and also to further compare the proposed framework with other RAG frameworks.

\normalem
\bibliographystyle{IEEEtran}
\bibliography{Reference}

\end{document}